\documentclass[conference]{IEEEtran}
\ifCLASSINFOpdf
   \usepackage[pdftex]{graphicx}
\else
   \usepackage[dvips]{graphicx}
\fi

\begin{document}
%
\title{Object Detection of Satellite Images Using \\Multi-Channel Higher-order Local Autocorrelation}

\author{\IEEEauthorblockN{Kazuki Uehara, Hidenori Sakanashi, Hirokazu Nosato, \\Masahiro Murakawa, Hiroki Miyamoto, and Ryosuke Nakamura}
\IEEEauthorblockA{National Institute of Advanced Industrial Science and Technology (AIST)\\
1-1-1, Umezono, Tsukuba city, Ibaraki, Japan 305-8560\\
Email: k-uehara@aist.go.jp}
}


%


\maketitle

\begin{abstract}
The Earth observation satellites have been monitoring the earth's surface for a long time, and the images taken by the satellites contain large amounts of valuable data. However, it is extremely hard work to manually analyze such huge data. Thus, a method of automatic object detection is needed for satellite images to facilitate efficient data analyses. This paper describes a new image feature extended from higher-order local autocorrelation to the object detection of multispectral satellite images. The feature has been extended to extract spectral inter-relationships in addition to spatial relationships to fully exploit multispectral information. The results of experiments with object detection tasks conducted to evaluate the effectiveness of the proposed feature extension indicate that the feature realized a higher performance compared to existing methods. 
\end{abstract}


%
\IEEEpeerreviewmaketitle

\renewcommand{\thefootnote}{\fnsymbol{footnote}}
\footnote[0]{\copyright ~2017 IEEE. Personal use of this material is permitted. Permission from IEEE must be obtained for all other uses, in any current or future media, including reprinting/republishing this material for advertising or promotional purposes, creating new collective works, for resale or redistribution to servers or lists, or reuse of any copyrighted component of this work in other works.}
\renewcommand{\thefootnote}{\arabic{footnote}}

\section{Introduction}
Earth observation satellites have been observing changes on the earth's surface for a long time, and the collected data is utilized for various purposes, such as land-use planning, disaster support, and climate-change monitoring. 
However, as such data is continuously accumulated over a wide area, it is extremely hard work to manually analyze the data. 
Hence, techniques of automatic object-detection technique are required in order to facilitate the efficient analyses of such an enormous amount of data. 
In this study, we apply an image object-detection technique to the satellite images.

Satellites typically provide two types of images: a panchromatic image and multispectral images. 
The panchromatic image is a single tonal image that captures a wide wavelength range, including a large part of the visible spectrum with high spatial resolution. 
In contrast, multispectral images consist of several images that are captured using instruments that are sensitive to specific wavelengths with low spatial resolution. 
In general, objects have unique spectral reflection and absorption characteristics within these specific spectral bands. 
Multispectral images are used for obtaining spectral information which is invisible to the human eye. 
Moreover, image processing such as inter-channel operations is performed to emphasize the spectral characteristics to analyze satellite images using multispectral images. 

For the object detection of satellite images, the panchromatic image is usually used because of its high spatial resolution \cite{wang11} \cite{bai14} \cite{zhang15}. 
In contrast, Newsam and Kamath have conducted image retrieval using typical texture features, such as GLCM feature and Gabor feature, on multispectral satellite images \cite{newsam04}. Their results indicate that utilizing all the multispectral images taken within the visible and near-infrared bands yielded a higher performance compared to only using a panchromatic image. 
However, given that such features were simply connected features extracted from the respective channels, it is also necessary to take into account the relationships among channels to exploit spectral information. 


In this paper, we therefore propose an image feature that considers the relationships among the channels, and evaluate the effectiveness of this feature within object-detection task experiments on satellite images. 
We focus on the higher-order local autocorrelation features (HLAC) \cite{otsu88} from the perspectives of its both calculation efficiency and recognition accuracy. 
HLAC is fast to compute because the feature-extraction procedure only involves product-sum operations. Thus, the feature is effective for large scale images, such as satellite images.
The feature can capture the local geometric patterns of images. 
However, as HLAC is unable to extract the relationships among channels,  
we have extended it to extract those relationships in addition to the local geometric patterns.

\section{Satellite Image}

%
Earth observation satellites are operated for monitoring of earth's surface. 
Their spatial and spectral resolutions depend on purpose of applications. 
For example, GeoEye-1 satellite provides high spatial resolution images with 0.41m, and it can capture four spectral images within the visible and the near-infrared bands \cite{geoeye}. 
SPOT-6 and SPOT-7 satellites can capture 1.5m spatial resolution of panchromatic image and four spectral images similar to the GeoEye-1 satellite \cite{spot}. 
These satellites are operated for commercial use. 

Landsat 8 is the latest satellite within the Landsat Project that has lasted over four decades and continues to observe the earth's surface with  a moderate spatial resolution of 15m to 100m \cite{roy14}.
It can capture multispectral images composed of 11 band images, including not only the visible light region, but also the near-infrared and the thermal-infrared regions, as shown in Table \ref{tab:sat_image}. 

\begin{table}[!t]
\renewcommand{\arraystretch}{1.3}
\caption{Spectral and spatial resolutions of Landsat 8 images.}
\label{tab:sat_image}
\centering
\begin{tabular}{l |c| c}
\hline
\bfseries Band & \bfseries Wavelength [$\mu$m] & \bfseries Resolution [m]\\
\hline \hline
1 (Ultra blue) & 0.43-0.45 & 30\\
2 (Visible, blue) & 0.45-0.51 & 30\\
3 (Visible, green)& 0.53-0.59 & 30\\
4 (Visible, red)& 0.64-0.67 & 30\\
5 (Near-infrared)& 0.85-0.88 & 30\\
6 (Short wavelength infrared) & 1.57-1.65 & 30\\
7 (Short wavelength infrared) & 2.11-2.29 & 30\\
8 (Panchromatic) & 0.50-0.68 & 15\\
9 (Cirrus) & 1.39-1.38 & 30\\
10 (Long wavelength infrared) & 10.60-11.19 & 100\\
11 (Long wavelength infrared) & 11.50-12.51 & 100\\
\hline
\end{tabular}
\end{table}

In this study, we use Landsat 8 because it can capture relatively more number of spectral bands than other satellites. 
Moreover, the Landsat image data is freely available \cite{landsat8}, and anyone can obtain the data about two hours after the satellite has taken the images. Thus, Landsat images are useful from the perspective of practical applications.

\section{Proposed Method}
\subsection{Higher-order local autocorrelation}
Higher-order local autocorrelation (HLAC) features can express local geometric patterns and are robust against noise. Moreover, the feature has important characteristic relating to shift invariance and additivity. The shift invariant property makes the same feature wherever an object is within an image. Additivity is the property whereby the overall feature for multiple objects is equal to the sum of the features for each object. As location and the number of objects within an image are unspecified, these properties are desirable for object-detection tasks. 

HLAC feature has been successfully applied to various applications, because of its calculation efficiency and recognition accuracy \cite{nanri05} \cite{iwata07} \cite{otsu11}.
The feature is calculated by adding the product of the intensity of a reference point $r=(x,y)$ and  predefined displacement vectors $a = (\Delta x, \Delta y)$ within local neighbors.  
The $N$th order of the HLAC feature is formulated by (1). 

\begin{eqnarray}
X_N(a_1, \cdots, a_N) = \sum f(r)f(r+a_1) \cdots f(r+a_N)
\end{eqnarray}
\noindent
where, $f (r)$ represents the gray level at the position $r$ of the image. The number of expressible features $X_N$ increases as its order increases, while its calculation cost also increases according to the order. Therefore, in calculations for practical use, the order is usually limited up to the second order. 

\begin{figure}[!t]
\centering
\includegraphics[width=8.5cm,bb=0 0 660 490]{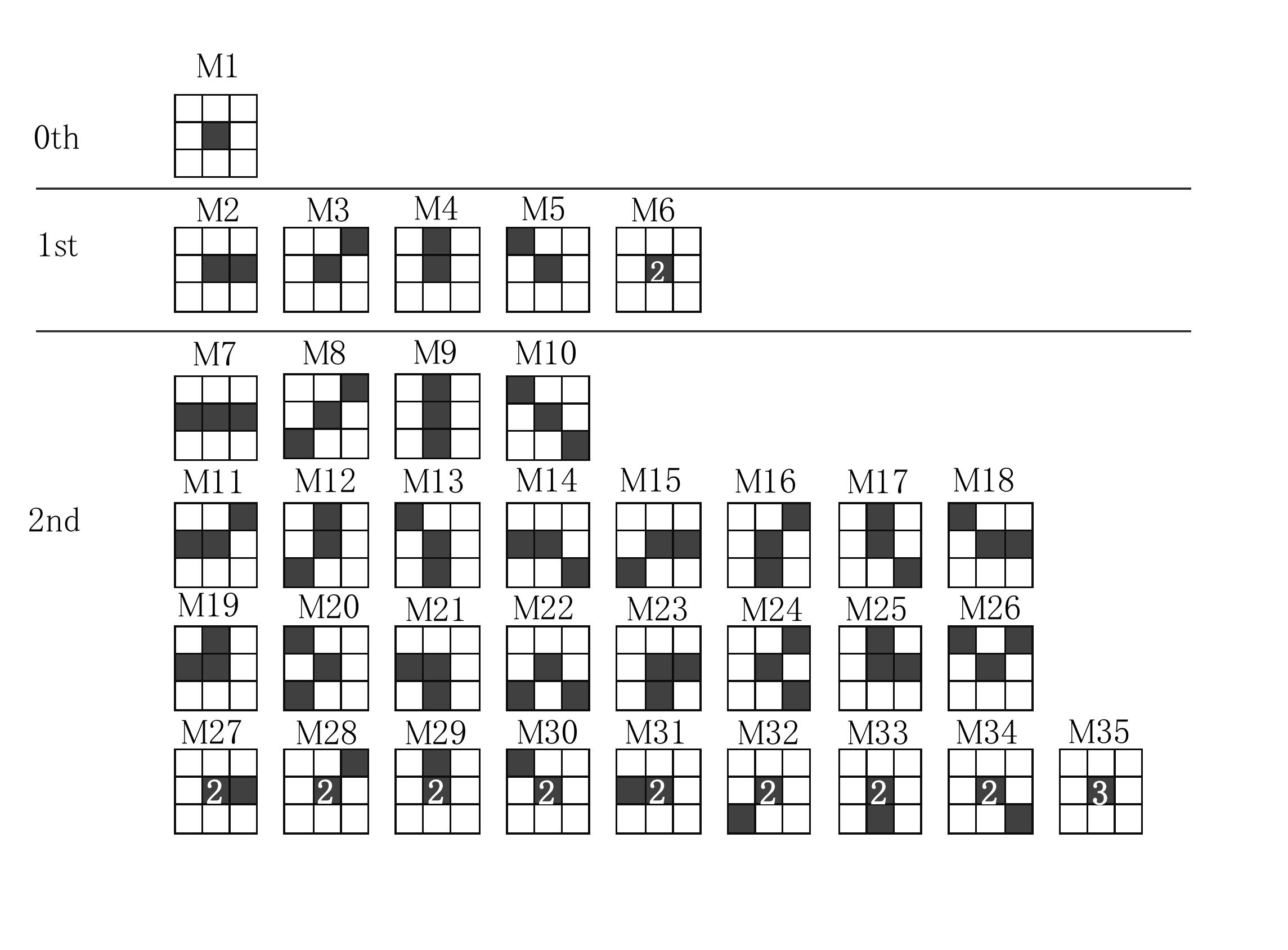}
\caption{HLAC mask patterns when local neighbors are 3 $\times$ 3 and order is limited to the second-order. }
\label{fig:hlac}
\end{figure}

Figure \ref{fig:hlac} shows 35 patterns of HLAC masks whose local neighbors are 3 $\times$ 3.  
In this case, patterns that are considered to be the same by shift are excluded. 
In each mask, ``black" represents the reference points in multiplying the values of the pixels, while ``white" represents points ``not required" for feature extraction. 
The numbers shown within some of the masks indicate the duplicated production of the gray level at the reference point. 

 
In order to extract global geometric features in addition to local geometric features, the size of local neighbors can be extended by the distance between the reference point and displacement points. The size of local neighbor is defined by $(2m+1)\times(2m+1)$, where $m$ (displacement distance) is an integer that is one or more. 
We combine multiple set of features extracted from the different size of displacement distances.

\subsection{Multi channel extension of HLAC}

\begin{figure}[!t]
\centering
\includegraphics[width=8cm,bb=0 0 505 755]{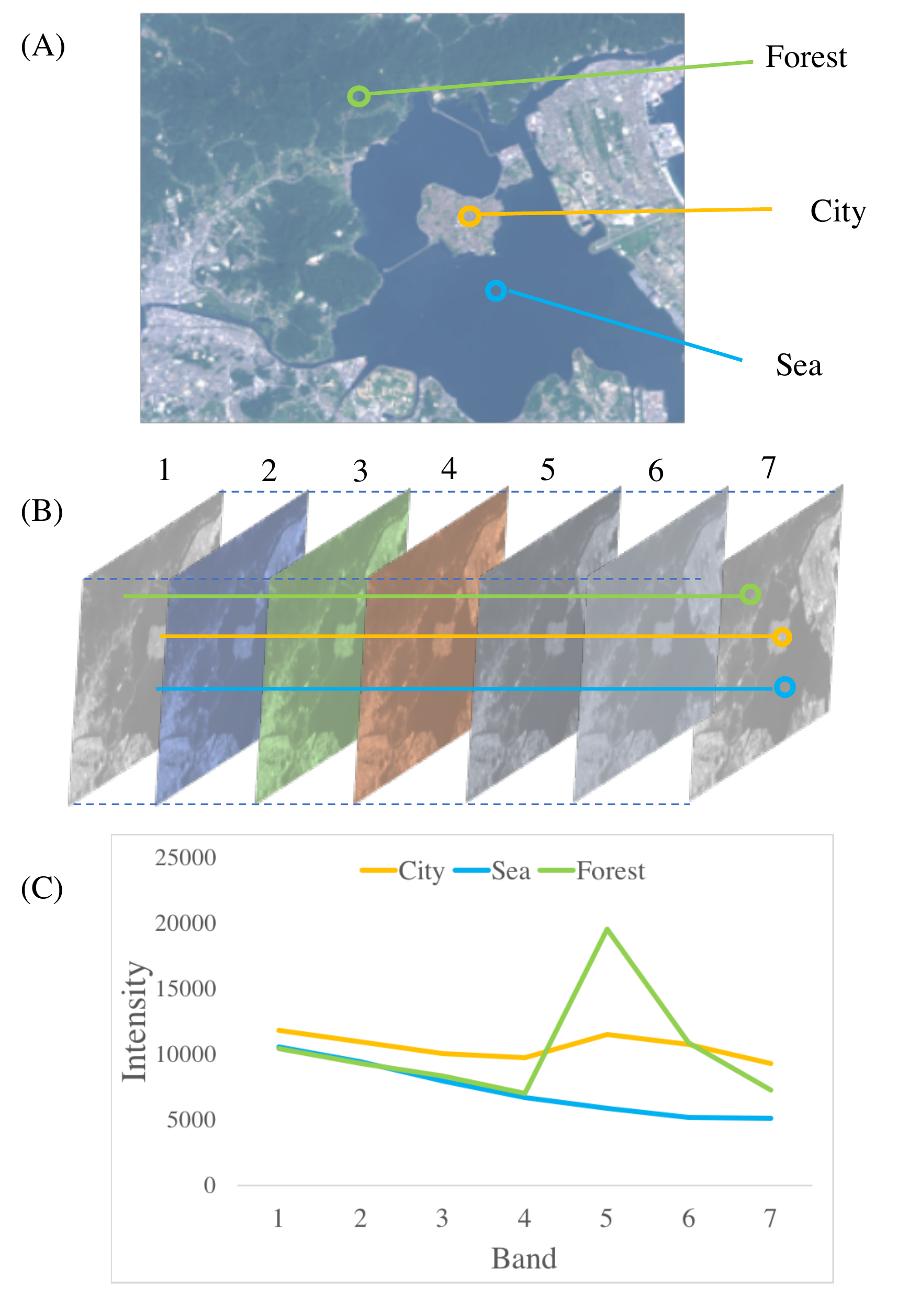}
\caption{(A): A color image made from visible (blue, green, red)  images. (B): Multispectral images taken at each band. (C): The differences in the spectral information for the three examples of forest, city, and sea. }
\label{fig:multi_spectral}
\end{figure}

Each channel of satellite image is obtained by observing the corresponding band of reflected light. 
In general, objects have unique spectral reflection and absorption characteristics, as shown in Fig. \ref{fig:multi_spectral}. 
For analysis of remote sensing images, the characteristics are usually emphasized by inter-channel operations. 
For example, the NDVI (Normalized Difference Vegetation Index), that is used as an indicator for the distribution and activity of vegetation, is based on the characteristic that plants absorb light in the red and reflect it in the near-infrared. The index is calculated based on the differences between two channels captured in the near-infrared and the red regions. The inter-channel operation makes the characteristic noticeable. 
As a result, these indices with higher value indicate dense vegetation area. 
In this way, the relationships among channels provide more information than obtainable from the channels independently. 



Thus, we have explicitly extended the HLAC feature to extract the relationship among channels (Multi Channel HLAC: MUCHLAC). With this modification, MUCHLAC can extract not only the spatial relationships, but also the spectral relationships of reference points. In this way, the feature is capable of extracting the complicated patterns among the spectral bands. 

More specifically, for first or higher orders of the HLAC, the MUCHLAC feature is computed by referring to two (or more) points from different channels. 
Then, the product of the reference points and the displacement points are summed by scanning the entire image.

\begin{figure}[!t]

\centering
\includegraphics[width=8cm,bb=0 0 425 250]{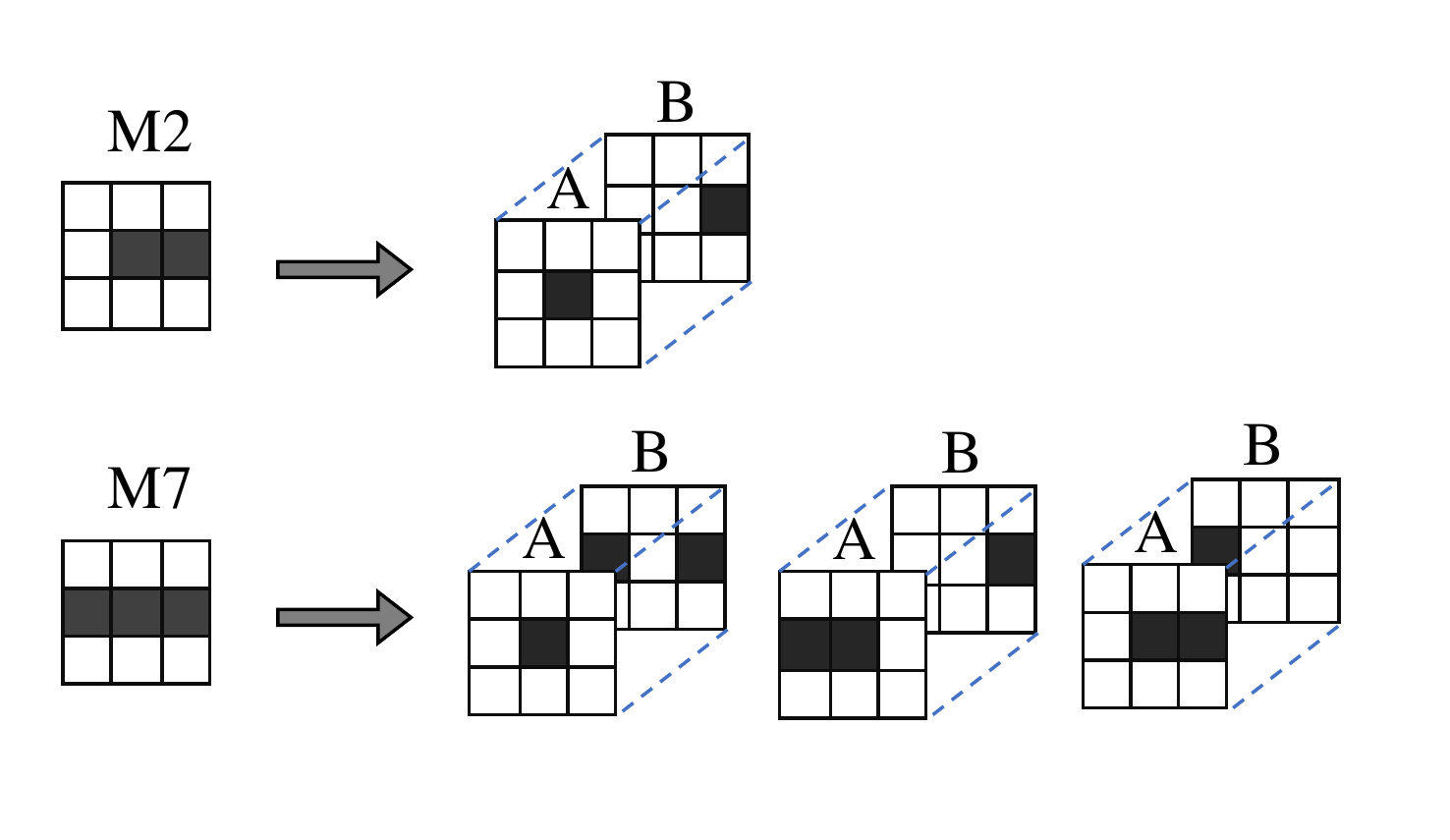}
\caption{Examples of the MUCHLAC mask patterns for first and second orders. Each mask pattern of the MUCHLAC is derived from each HLAC mask pattern. }
\label{fig:muchlac}
\end{figure}

Figure \ref{fig:muchlac} presents examples of MUCHLAC mask patterns combining two channels (A, B) when extracting first and second order features. 
The displacement point with respect to the reference point of channel A is selected from either A or B. Here, at least one displacement point must be selected from a different channel from the one containing the reference point. 
Feature extraction for the other masks are the same as these examples. 
The number of mask patterns for MUCHLAC is five patterns in the first order and 77 patterns in the second order. 
Mask patterns that are equivalent in terms of shift and channel replacement are eliminated.

In the case of extracting features by combining $n$ channels from a multi-channel image consisting of $M$ channels, feature extraction is performed from all  pairs, including the permutations of these $n$ channels. Consequently, the number of combinations is $_MP_n$.
For example, MUCHLAC feature extraction from a typical color image consisting of 3 channels (RGB) is as follows: 
At first, the HLAC feature is extracted from each channel, respectively, and they are connected in series. 
Then, the MUCHLAC feature is extracted from each pair of channels considering all permutation, and it is then connected with the HLAC feature.  
Consequently, the length of the feature vector is 597(35 $\times$ 3 + 82$\times$6). 


\section{Experiment}
In order to evaluate the effectiveness of the proposed feature, object-detection tasks were conducted on satellite images using both the HLAC feature, and the MUCHLAC feature. In addition, we tested the GLCM (gray level co-occurrence matrix) feature on this recognition task, because it is widely used for remote sensing images and has been shown to exhibit higher performance in \cite{newsam04}.

\begin{figure}[!t]
\centering
\includegraphics[width=8.5cm,bb=0 0 715 475]{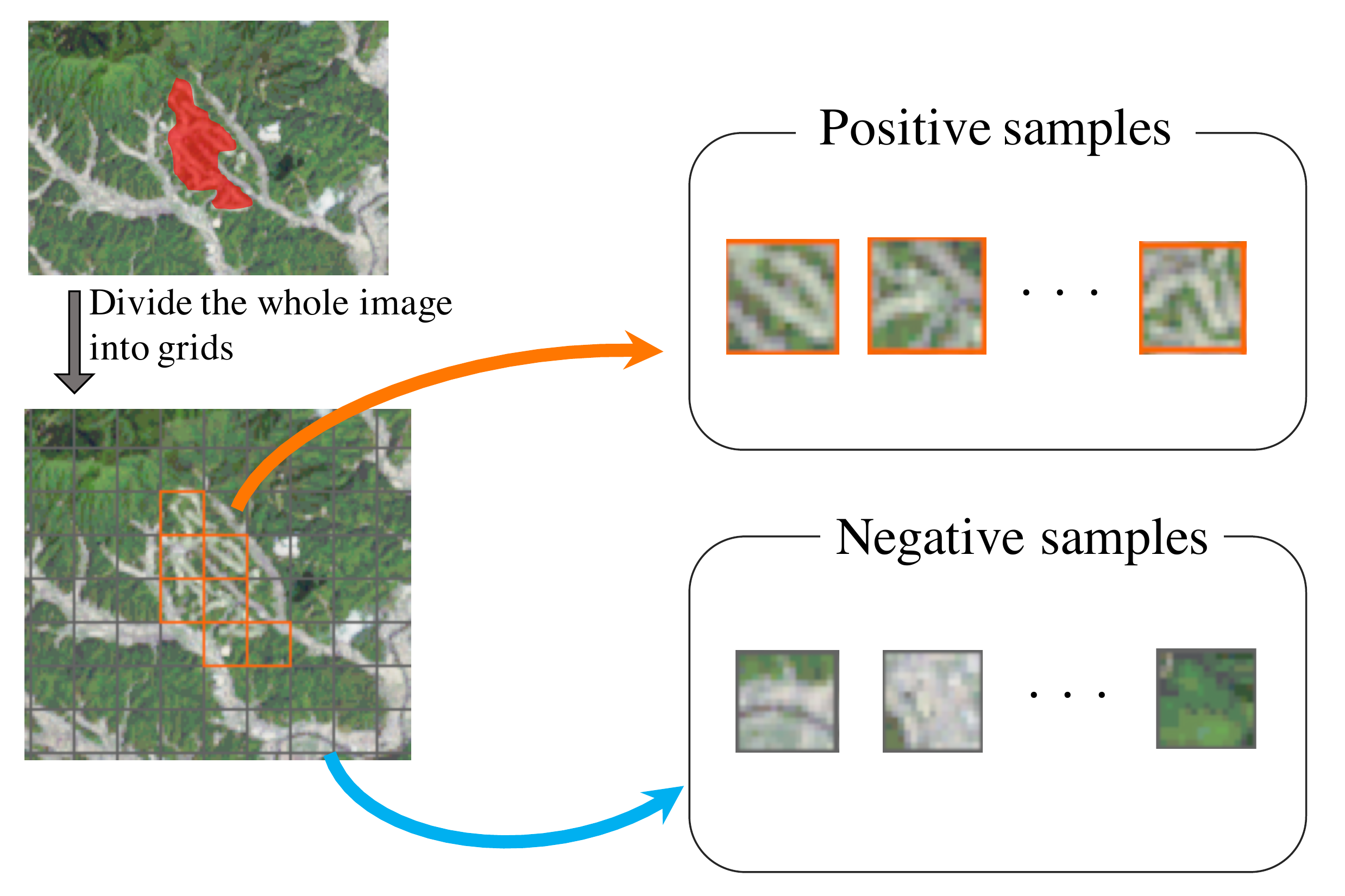}
\caption{An illustration of how the datasets are created. The complete image is divided into small patches. An image that includes part of a golf course is a positive sample, while one that does not is a negative sample. }
\label{fig:sampling}
\end{figure}

\begin{figure*}[!t]
\centering
\includegraphics[width=18cm,bb=0 0 1233 308]{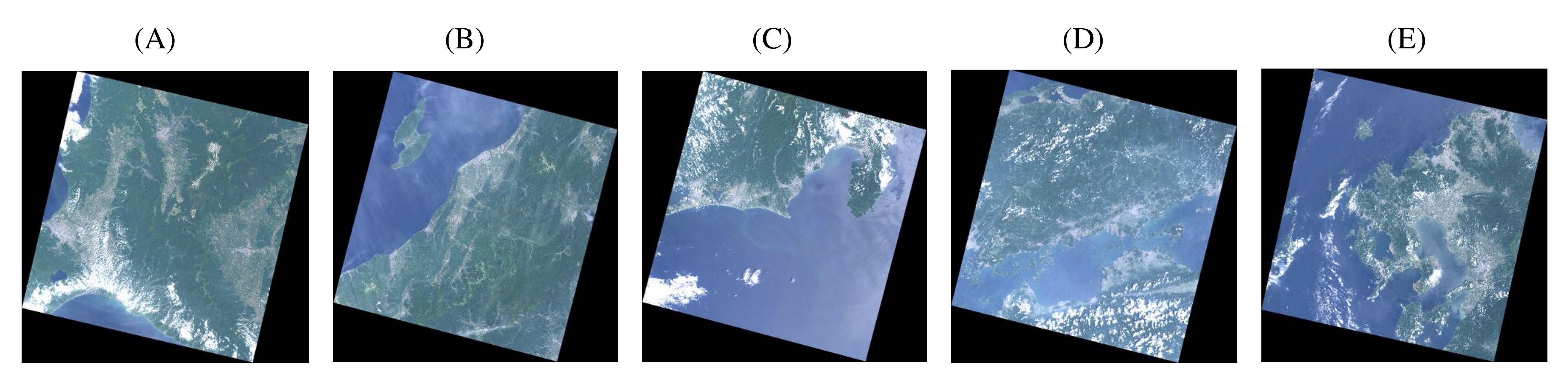}
\caption{The satellite images used in the experiments. The ID numbers for each satellite image are as follows, (A): LC81070302016189, (B): LC81080342015193, (C): LC81080362016196, (D): LC81110362015214, (E): LC81130372015212.}
\label{fig:satellite}
\end{figure*}

\subsection{Dataset}
A satellite image is divided into a grid of patch images. The task is to recognize whether each patch image includes target object or not. We chose golf courses as the target for the task, because it encompasses considerable variations, in terms of sizes, course arrangements, and the vegetation present. The size of an image patch is 16 $\times$ 16 pixel. Images that include a part of a golf course are defined as positive samples, while those that do not are defined as negative samples (Fig. \ref{fig:sampling}).

In order to create training and evaluation samples, we selected five satellite images for a part of Japan that were taken in either the July or August of 2015 and 2016 with relatively few clouds (Fig. \ref{fig:satellite}). 
The details of the satellite images are shown in Table \ref{tab:experiment_data}. 
The size of each image is about 7800 $\times$ 8000 pixels, and each image covers approximately 170km $\times$ 185km. 

\begin{table}[!t]
\renewcommand{\arraystretch}{1.3}
\caption{The number of samples used in the experiments.}
\label{tab:experiment_data}
\centering
\begin{tabular}{c |c |c}
\hline
\bfseries Image ID number & \bfseries \# of positives & \bfseries \# of negatives\\
\hline \hline
LC81070302016189 & 610 & 16000 \\
LC81080342015193 & 676 & 16000 \\
LC81080362016196 & 570 & 16000 \\
LC81110362015214 & 897 & 16000 \\
LC81130372015212 & 697 & 16000 \\
\hline
\end{tabular}
\end{table}

\subsection{Image Features}

\subsubsection{MUCHLAC}
In MUCHLAC feature extraction, features are extracted for each channel and each displacement distance. Furthermore, the feature extracted by combinations between channels is concatenated. 
HLAC feature can be reconstructed for rotation and reflection invariance \cite{nosato11}.
The MUCHLAC feature also can be reconstructed in an identical manner to that of the HLAC. 
We reconstructed the feature for rotation and reflection invariance. 
In order to exploit the spectral information, the feature was extracted from seven bands, which were 1, 2, 3, 4, 5, 6, and 7. 
The displacement distances used for feature extraction were from 1 to 4. 
The distances were determined by preliminary experiments\footnote{The preliminary experiments were carried out in order to determine the parameters of the features and the classifier. We investigated several different parameters applied to the object detection tasks for satellite images that were different from the images shown in Table\ref{tab:experiment_data}, with respect to both their locations and dates.  } that yielded high detection performances of both HLAC and MUCHLAC. 
In addition, the number of channel combinations was set to two. 


\subsubsection{HLAC}
In HLAC feature extraction, features are calculated for each channel and each displacement distance from the mask patterns, and connected sequentially. 
After extracting the HLAC, this feature is also reconstructed for rotation and reflection invariance. 
The HLAC feature was extracted from seven bands and from four displacement distances from 1 to 4 similar to the MUCHLAC. 

\subsubsection{Gray Level Co-occurrence Matrices}
Texture features based on the spatial dependence of pixel values \cite{haralick73} are popular for the analysis of remote sensing images. 
The feature is calculated using GLCM that tabulate how often different combinations of gray levels occur within an image. 
Originally, 14 quantities were proposed as the feature, but, typically, only a subset of the quantities are used. 
We chose the five quantities of {\it angular second moment}, {\it contrast}, {\it inverse different moment}, {\it entropy}, and {\it correlation}, which are the same as \cite{newsam04}. 
Moreover, the angles for calculating the GLCMs were 0$^\circ$, 45$^\circ$, 90$^\circ$, 135$^\circ$.
The feature was extracted from seven bands the same as both MUCHLAC and HLAC. 


\subsection{Classifier}
We used Real AdaBoost \cite{schapire99} for the classification of the patch images. 
We set the number of weak learners based on the preliminary experiment in order to enhance recognition performance. 

\subsection{Evaluation Criterion}
We chose precision, recall, and the F-measure as evaluation criteria. 
Precision is the fraction of retrieved instances that are relevant (correctness), while recall is the fraction of relevant instances that are retrieved (completeness). 
The F-measure is a weighted harmonic mean of precision and recall, with a non-negative value for weight (2). 
\begin{eqnarray}
F \mbox{-} measure &=& \frac{(\beta^2+1) \cdot precision \cdot recall}{\beta^2 \cdot precision + recall}
\end{eqnarray}

As $\beta$ is commonly set to 1, we set it to 1. 

\subsection{Results and Discussion}
\subsubsection{Detection Results}\label{sec:result}
Table \ref{tab:detection_result} presents the averages for the detection results, which were analyzed by five-fold cross validation (CV). 
In the table, the abbreviation TP (True Positive) is the number of samples correctly predicted as positive, FP (False Positive) is the number of negative samples that are incorrectly classified as positive. Similar to TP and FP, respectively, TN (True Negative) is the number of samples correctly predicted as negative, and FN (False Negative) is the number of positive samples that are incorrectly classified as negative. 

The result indicate that MUCHLAC exhibited higher performance than other features, in terms of all evaluation metrics. 
The performance levels of the HLAC and the GLCM were almost the same in this experiment.

\begin{table}[!t]
\renewcommand{\arraystretch}{1.3}
\caption{Detection results for patch image classification.}
\label{tab:detection_result}
\centering
\begin{tabular}{l |c |c|c}
\hline
   & \bfseries MUCHLAC (ours) & \bfseries  HLAC & \bfseries GLCM\\
\hline \hline
TP & 516.8 & 479.2 & 478.2 \\
FP & 62.2 & 90.4 & 90.6 \\
TN & 15934 & 15905.8 & 15905.6 \\
FN & 172.2 & 209.8 & 210.8 \\ \hline \hline
Precision & {\bf 0.89} & 0.84 & 0.84 \\
Recall & {\bf 0.75} & 0.7 & 0.69\\
F measure & {\bf 0.82} & 0.76 & 0.76\\
\hline
\end{tabular}
\end{table}

Each component of the MUCHLAC feature is calculated from each channel and combinations of multiple channels. 
Thus, the MUCHLAC contains all the HLAC components. 

In order to verify the contribution of the feature extracted from the combinations of the multiple channels to object detection, we calculated the importance of each feature component of the MUCHLAC using a Random Forest \cite{breiman01} \cite{louppe13}. 

The calculation of feature importance is based on the idea that, if the feature is unimportant, permuting that component will not reduce prediction accuracy. 
To calculate the importance of a given feature component, the values for the component are randomly permuted within out-of-bag samples that are unused for forest construction. Then,  the difference in terms of prediction accuracies between before and after permutation can be used as an index of importance.

Focusing on the importance of the top 100 feature components, the feature component extracted from the combinations of multiple channels occupied 90\%. 
From both this feature component analysis and the experimental results, clearly, feature extraction that considers the interrelationships among multiple channels is effective for the object detection of multispectral images, 

\subsubsection{The number of training samples}

\begin{figure*}[!t]
\centering
\includegraphics[width=11cm,bb=0 0 532 257]{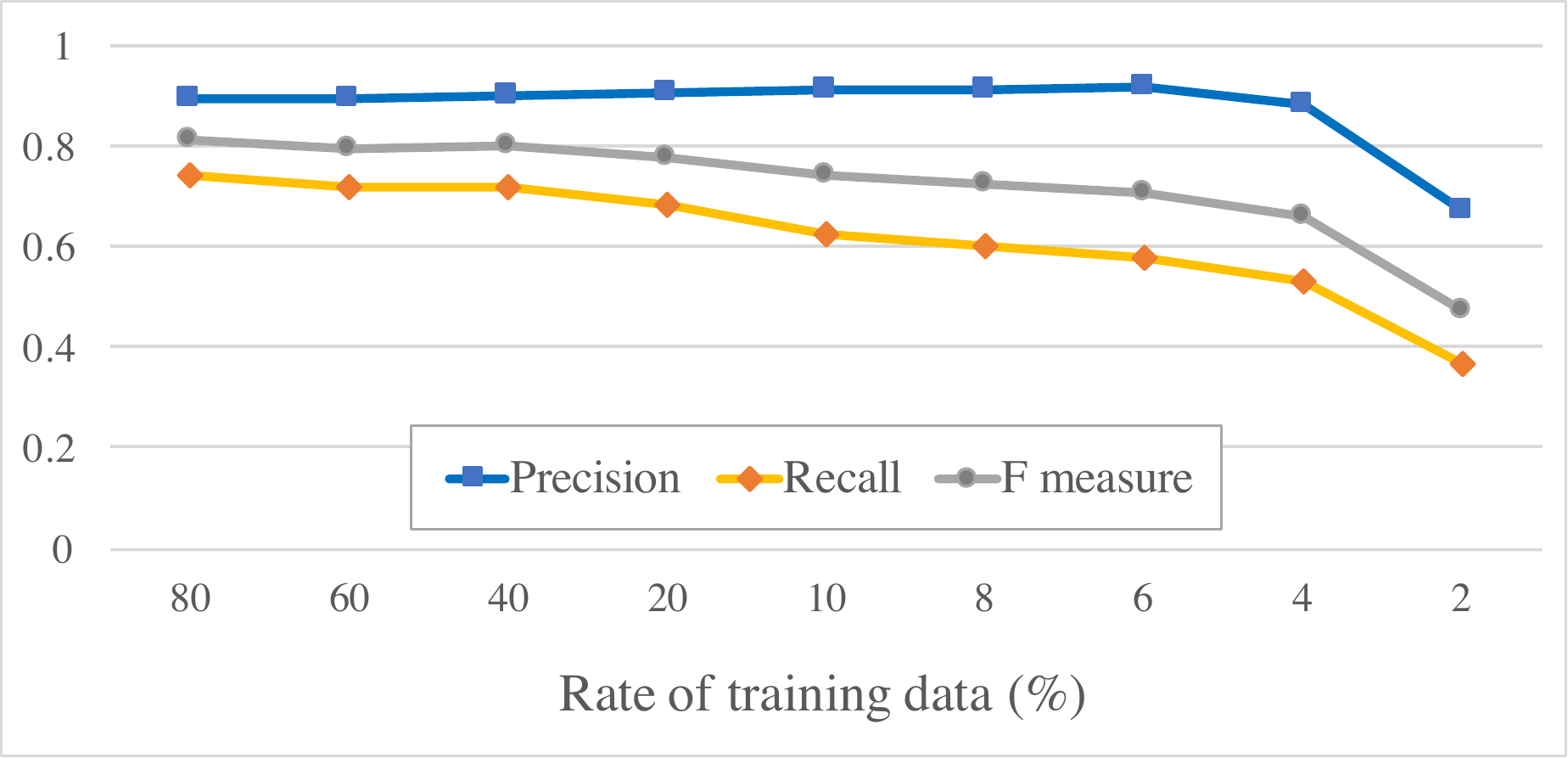}
\caption{Detection accuracy results as a function of reducing the training data. }
\label{fig:sensitivity}
\end{figure*}

\begin{table*}[!t]
\renewcommand{\arraystretch}{1.3}
\caption{The size of the training data sets used in the sensitivity experiment. }
\label{tab:sensitivity}
\centering
\begin{tabular}{l |c|c|c|c|c |c| c|c|c}
\hline
Ratio (\%) & 80 & 60 & 40 & 20 & 10 & 8 & 6 & 4 & 2 \\  \hline \hline
The number of positive samples & 2223.6 &1675 &1129.8 & 565.6 & 276.2 & 232.2 & 172.2 & 120.8 & 54.2\\
The number of negative samples & 51199.2 & 38300.6 & 25612.6 & 12814.8 & 6387.2 & 5127 & 3879 & 2497.8 & 1055.2\\ \hline
Total & 53422.8 & 39975.6 & 26742.4 & 13380.4 & 6663.4 & 5359.2 & 4051.2 & 2618.6 &1109.4 \\ \hline
\end{tabular}
\end{table*}

For remote sensing images, the set of labeled data is limited and labeling task is very time consuming due to the vast amount of data. 
Accordingly, it is desirable to achieve high accuracy with small amounts of data. 
In this study, we also investigated detection performance under scenarios where the number of training samples is reduced by some ratio. 
The data used for evaluation is the same as that of the previous experiment. 
In this experiment, five-fold CV was carried out for each ratio of the data set. 

Table \ref{tab:sensitivity} shows the average sizes of the training samples used in the experiment, and 
Fig. \ref{fig:sensitivity} presents the results of the experiment. 

As Fig. \ref{fig:sensitivity} illustrates, performance improved as the number of samples increased. 
Performance sharply increases between 2\% (about 1100 samples) and 6\% (about 4000 samples). 
For conditions with more than 6\% of the data, the performance in terms of F-measure exceeded 0.7. 
Considering the number of positive samples which is about 170 at the 6\% ratio, the performance is reasonable compared with the results achieved by using the complete set of training samples.
This result indicates the system can achieve relatively high performance even with a small number of samples. 



\subsubsection{The number of spectral bands}

\begin{figure}[!t]
\centering
\includegraphics[width=8.5cm,bb=0 0 541 323]{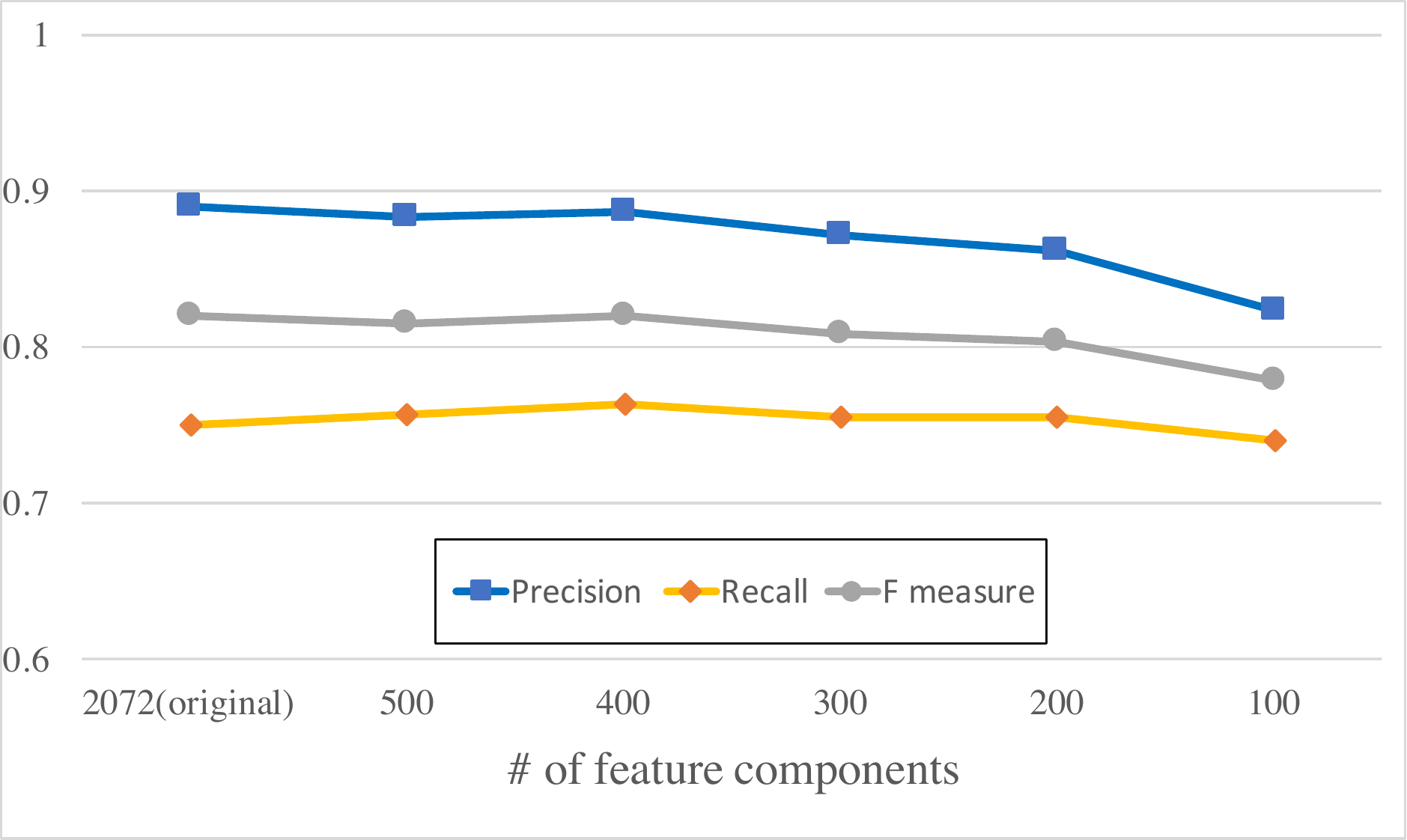}
\caption{Detection performance results for the MUCHLAC as a function of number of features selected. }
\label{fig:dim_reduce}
\end{figure}

In the case of images obtained from satellites equipped sensors with higher spectral resolution, it is possible to extract even more complicated interrelationships among the channels. 
However, as the number of channels increases, naturally, the number of feature dimensions also increases, which leads to increases in computational times. 
Thus, it is important to consider which feature components to select depending on the problem being tackled. 

This sub-section reports on an additional experiment conducted to evaluate an approach to feature selection that balances reductions in the dimensions of the MUCHLAC while maintaining performance. 
The selection procedure is as follows. 
First, the importance of each component is calculated using a Random Forest. 
Then, the components are selected in descending order of their importance. 

Recognition performances for conditions of selecting between 100 to 500 components in 100 intervals are shown in Fig. \ref{fig:dim_reduce}. 
In this experiment, the train and the test samples are the same as those described in section \ref{sec:result}.

The figure indicates that performance drops slightly as the number of feature components is decrease. 
The levels of performance are constant when the number of dimensions has more than 400 components. 
This result suggests that most of the feature components are redundant for the detection task. 
Thus, it is practical to reduce the dimensions of the feature based on the importance of the feature components. 

\section{Conclusion}
In this paper, we have proposed a new image feature that extends on the higher order local autocorrelation feature to extract not only spatial relationships but also spectral relationships from multi spectral satellite images. The feature is able to exploit multispectral information. 
In order to evaluate the effectiveness of the feature extraction, we conducted object detection task on satellite images and compared the proposed feature with existing features. 
As a result, the proposed feature yielded higher levels of performance than the conventional features. 
Moreover, we investigated a feature analysis based on the feature importance using a random forest. 
These results indicate that the feature extraction taking into account relationships among channels is effective for object detection of multispectral satellite images. 



\section*{Acknowledgment}
This paper is based on results obtained from a project commissioned by the New Energy and Industrial Technology Development Organization (NEDO).



%

\end{document}